%% file: main.tex
\documentclass[11pt,a4paper]{article} 
\usepackage[hyperref]{emnlp2020}
\usepackage{times}
\usepackage{latexsym}
\usepackage{graphicx}

\usepackage[linesnumbered,ruled,vlined]{algorithm2e}
\usepackage{microtype}
\usepackage{amsfonts,amssymb}
\usepackage{amsmath}
\usepackage{booktabs}
\usepackage{multirow}
\usepackage{url}
\usepackage{cleveref}
\usepackage{subcaption, caption}
\makeatletter
\g@addto@macro{\UrlBreaks}{\UrlOrds}
\makeatother

\aclfinalcopy 

\title{Learning a Simple and Effective Model for Multi-turn Response Generation with Auxiliary Tasks }

\author{
Yufan Zhao$^1$, Can Xu$^1$\thanks{Corresponding author: Can Xu (caxu@microsoft.com).}~,
Wei Wu$^2$,
Lei Yu$^3$ \\
$^1$Microsoft Corporation, Beijing, China\\
$^2$Meituan, Beijing, China \\
$^3$Beihang University, Beijing, China \\
\texttt{\{yufzhao,caxu\}@microsoft.com} \\
\texttt{wuwei19850318@gmail.com}\\
\texttt{yulei@buaa.edu.cn}
}

\date{}

\begin{document}
\maketitle

\begin{abstract}
We study multi-turn response generation for open-domain dialogues. The existing state-of-the-art addresses the problem with deep neural architectures. While these models improved response quality, their complexity also hinders the application of the models in real systems.  In this work, we pursue a model that has a simple structure yet can effectively leverage conversation contexts for response generation. To this end, we propose four auxiliary tasks including word order recovery, utterance order recovery, masked word recovery, and masked utterance recovery, and optimize the objectives of these tasks together with maximizing the likelihood of generation. By this means, the auxiliary tasks that relate to context understanding can guide the learning of the generation model to achieve a better local optimum. Empirical studies with three benchmarks indicate that our model can significantly outperform state-of-the-art generation models in terms of response quality on both automatic evaluation and human judgment, and at the same time enjoys a much faster decoding process.   
\end{abstract}
       \input{introduction}
	   \input{related_work}	 
	   \input{approach}
	   \input{experiment}
 	   \input{conclusion}
\newpage

\bibliography{emnlp2020}
\bibliographystyle{acl_natbib}
\appendix
\input{Supplementary}
\end{document}

%% file: introduction.tex
\section{Introduction}
As an important topic in conversational AI, open-domain human-machine conversation is gaining increasing attention from both academia and industry. A common approach to building such a system is to learn a response generation model within an encoder-decoder framework using neural sequence architectures \cite{sutskever2014sequence,vaswani2017attention}. While the encoder-decoder framework has been successfully applied in various text generation tasks such as machine translation \cite{vaswani2017attention}, summarization \cite{rush2015neural}, paraphrase generation \cite{dong2017learning}, etc., it has to deal with a unique challenge in the task of response generation: modeling conversation contexts. A conversation context often exhibits a hierarchical structure with dependency existing on both a word-level and an utterance-level. Moreover, as indicated in \cite{xing2017hierarchical,zhang2019recosa}, information in a context is rather redundant for responding: commonly only a few words and utterances are useful for response generation, and the positions of the relevant words and utterances vary from case to case. To model the hierarchy of conversation contexts, hierarchical recurrent encoder-decoder (HRED) \cite{serban2016building} extends the vanilla sequence-to-sequence model by a word-level encoder and an utterance-level encoder. Later on, a hierarchical recurrent attention network  (HRAN) \cite{xing2017hierarchical} harnesses the decoder of the HRED model with word-level attention and utterance-level attention to dynamically highlight the effect of relevant words and utterances in response synthesis. Very recently, ReCoSa \cite{zhang2019recosa} further exploits multi-layer multi-head self-attention\footnote{The fact that both the encoder and the decoder of ReCoSa contain multiple layers is not highlighted in the paper, but is revealed by the source code released by the authors at \url{https://github.com/zhanghainan/ReCoSa}.} to model long-term dependency among utterances and responses. From HRED to HRAN, and then to ReCoSa, the performance of the models in terms of response quality becomes better and better \cite{zhang2019recosa}, but the models also grow to be more and more complicated. For example, the number of parameters in ReCoSa is more than twice as that in HRED. Thus, when we enjoy the improved performance from the increased complexity, the complexity may also impede the application of the models in some scenarios (e.g., in a mobile scenario).

In this work, we study multi-turn response generation and target on a model that has a simple structure yet can make use of conversation contexts as well as the existing deep models. The key idea is to transfer the burden of context understanding from modeling to learning by designing several auxiliary tasks, and leverage the auxiliary tasks as regularization in model estimation. Specifically, the model we use for response generation concatenates utterances in a conversation context as a long sequence, and only exploits one-layer self-attention in encoding and one-layer context attention in decoding. In such a frugal setting, the representation capability of the model shrinks a lot compared with deep transformers. As a remedy, we augment the maximum likelihood estimation (MLE) in learning with objectives from four auxiliary tasks including word order recovery, utterance order recovery, masked word recovery, and masked utterance recovery.  In the first two tasks, we predict the correct order of words and utterances from a random shuffle of words in an utterance and a random shuffle of utterances in a context respectively. The goal of the two tasks is to enhance understanding of the sequential dependency among words and utterances within a context. The other two tasks are inspired by the recent breakthrough from BERT \cite{devlin2019bert}, in which we randomly mask a word in an utterance and an utterance in a context respectively, and predict the masked word and the masked utterance using the remaining words and utterances. The two tasks may encourage the learning process to pay more attention to semantics of words and utterances in their contexts, and help the learning process find better representations of words and utterances for the generation model. The auxiliary tasks and the MLE task share the encoder of the generation model. Through learning with multiple tasks, optimization for response generation and optimization for context understanding are performed in a joint form. The context understanding related tasks can guide the MLE to achieve a better local optimum, and thus realize superior performance in response generation with a simple neural structure. 

We test the proposed approach with three benchmarks including the Ubuntu Dialogue Corpus \cite{lowe2015ubuntu}, DailyDialog \cite{li2017dailydialog}, and PERSONA-CHAT \cite{zhang2018personalizing}. Evaluation results on all  three datasets indicate that our model can significantly outperform state-of-the-art generation models in terms of both automatic evaluation and human judgment. Moreover, with a parameter set even smaller than HRED, our model is 2x faster than ReCoSa in response decoding.

Our contributions in the paper are three-fold: (1) proposal of balancing model complexity and model capability in multi-turn response generation; (2) proposal of four auxiliary learning tasks that transfer context understanding from modeling to learning; and (3) empirical verification of the effectiveness and the efficiency of the proposed model on three benchmarks. 

%% file: related_work.tex
\section{Related Work}
End-to-end open-domain dialogue generation is built upon the encoder-decoder architecture \cite{shangL2015neural,vinyals2015neural}, and the vanilla sequence-to-sequence structure has been widely extended to address challenges such as generic responses \cite{li2015diversity,xing2017topic}, context modeling \cite{serban2016building,serban2017hierarchical,xing2017hierarchical,zhang2019recosa}, and grounding by persona/emotion/knowledge \cite{li2016persona,zhang2018personalizing,zhou2017emotional,dinan2018wizard}. In this work, we study how to leverage conversation context for multi-turn response generation, which represents a fundamental problem in dialogue generation.  Different from the existing work that enhances the representation capability of models through neural architecture engineering, we turn to an orthogonal direction that we keep the generation model simple, and optimize the simple structure by learning with auxiliary tasks that encode context understanding.  As a result,  our model can provide high-quality responses at a low cost. Before us, there have been a few studies on learning a primary task with auxiliary ones \cite{rei2017auxiliary,yu2016learning,ding2017recurrent,trinh2018learning,mehri2019pretraining,wu2019self}. The work is unique in that through extensive empirical studies, we verified that a simple structure learned with auxiliary tasks can work as well as deep architectures in dialogue generation. 

%% file: approach.tex
\section{Approach}
We first formalize the problem in question, and then detail the model and the learning tasks.
\subsection{Problem Formalization}
Suppose that we have a dataset $\mathcal{D}=\{(\mathcal{U}_i,R_i)\}^N_{i=1}$, where $\mathcal{U}_i$ = $(U_{i,1},\ldots,U_{i,n})$ denotes a context with $U_{i,j}$ the $j$-th utterance, and $R_i$ is a response regarding to $\mathcal{U}_i$. The goal is to estimate a generation probability distribution $P(R | \mathcal{U})$ from $\mathcal{D}$, and thus, given a new context $\mathcal{U}$, one can generate a response for $\mathcal{U}$ following $P(R | \mathcal{U})$. A common practice is to learn $P(R|\mathcal{U})$ by maximizing the log-likelihood of $\mathcal{D}$ (i.e. MLE) which can be formulated as
\begin{equation}\label{mle}
\sum_{i=1}^N \log P(R_i|\mathcal{U}_i).
\end{equation}
When $P(R|\mathcal{U})$ is in a simple structure, only learning with MLE could be insufficient to obtain a model that can well capture the syntax and the semantics of contexts. An evidence is that simple architectures like HRED is much worse than complicated architectures like ReCoSa in terms of response quality, as reported by the existing work \cite{zhang2019recosa}. Since a simple structure is still favored, we consider aiding the objective given by Equation (\ref{mle}) with extra ones that can reinforce context understanding in the learning process. 

\subsection{Generation Model}

\begin{figure}[t!]
    \centering
    \includegraphics[width=\columnwidth]{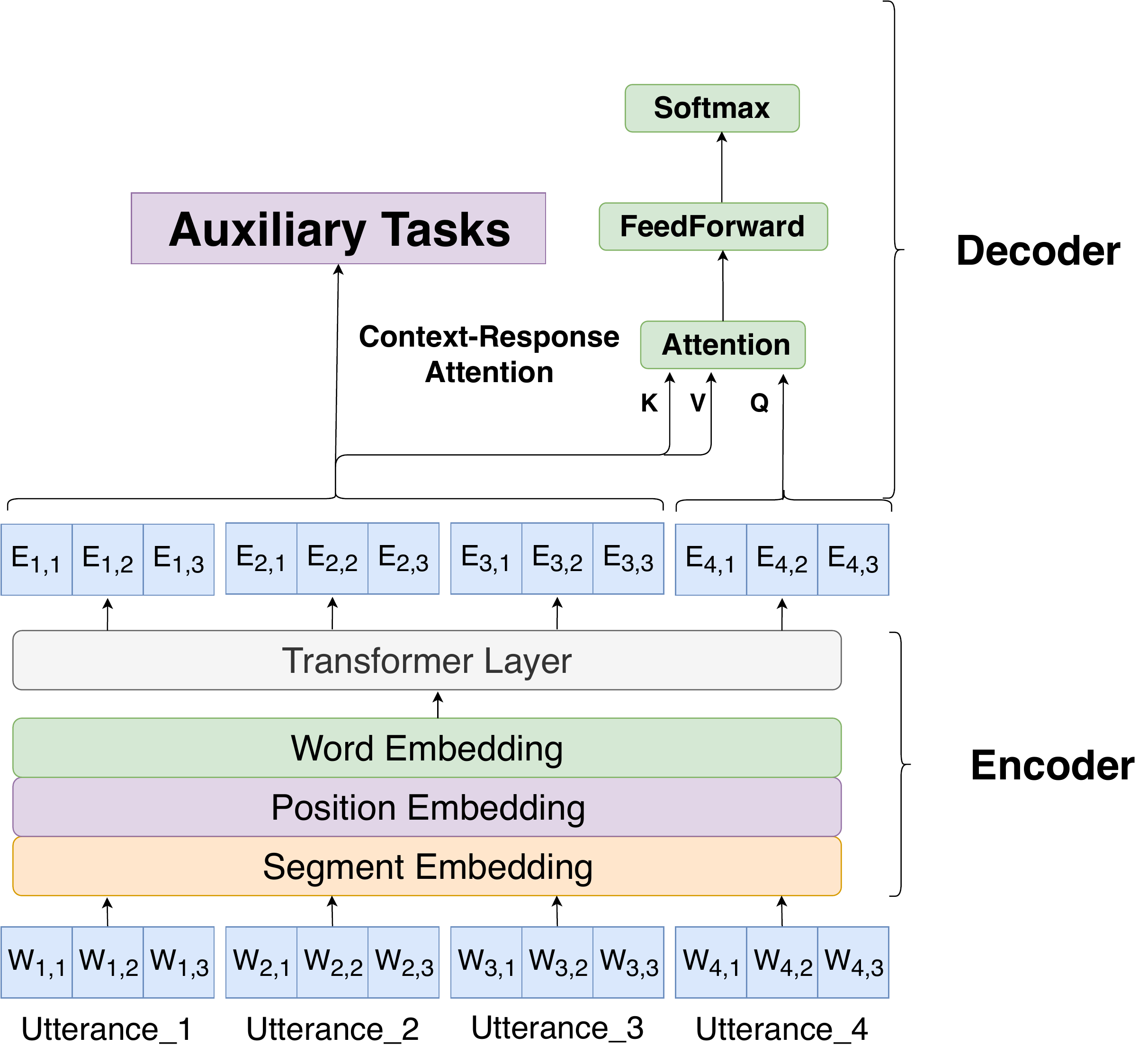}
    \caption{Architecture of the generation model.}
    \label{fig:model}
\end{figure}

Figure \ref{fig:model} illustrates the architecture of the generation model. In a nutshell, the model is in a transformer-based structure \cite{vaswani2017attention} with one attentive layer (in the transformer layer) in the encoder and one attentive layer in the decoder. The auxiliary tasks, which will be presented later, share the encoder with the generation model. We prefer a transformer-based structure instead of a recurrent structure, because the former is easier to parallelize than the latter, and thus can further enhance efficiency of the model in an online system. 
\paragraph{Encoder:}
we unfold all words in $(\mathcal{U},R)$ into $\mathcal{W}=(w_1,\dots,w_m,w_{m+1},\dots,w_{m+t})$, where $m$ is the number of words in context $\mathcal{U}$, and $t$ is the number of words in response $R$. 
$\forall i \in \{1,\ldots, m+t\}$, $w_i$ is represented by a summation of word embedding, position embedding, and segment embedding:
\begin{equation}\label{input_embed}
B(w_i) = WE(w_i) + PE(w_i) + SE(w_i),
\end{equation}
where $WE(w_i)$ represents the word embedding of $w_i$ initialized using GloVe \cite{pennington2014glove}, $PE(w_i)$ is the position embedding of $w_i$ which is defined by $P e(w_i)$, where $e(w_i)$ is a one-hot vector with the only non-zero entry indicating the position of $w_i$ in $\mathcal{W}$, and $P \in \mathbb{R}^{d \times M_p}$ is a randomly initialized matrix with $M_p$ an upper bound of the number of words in a dialogue. $SE(w_i)$ is the segment embedding of $w_i$ defined similarly with the one-hot vector indicating the position of the utterance that contains $w_i$. The embedding matrix is then fed to a transformer layer, which can be formulated as
\begin{equation}\label{multihead}
\begin{split}
& I = [B(w_1),B(w_2),\dots,B(w_{m+t})], \\
& E = \text{FNN}(\text{MultiHead}(I,I,I)),
\end{split}
\end{equation}
where $\text{FNN}(\cdot)$ is a feed-forward neural network and
$\text{MultiHead}(Q, K, V)$ is a multi-head attention
function with $Q$ a query, $K$ a key, and $V$ a value. To control the receptive field of self-attention in different tasks, we add a mask matrix $M \in \mathbb{R}^{(m+t) \times (m+t)}$ \cite{dong2019unified} in attention computation, and let $M$ determine whether a pair of words can attend to each other according to the learning tasks. Thus, $\text{MultiHead}(Q, K, V)$ is defined by
\begin{equation}\label{selfatt}
\begin{split}
&\text{MultiHead}(Q, K, V) =\oplus_{i=1}^K \text{Head}_i(Q,K,V),\\
&\text{Head}_i(Q,K,V) =\text{Attention}(W_iQ,W_iK,W_iV),\\
&\text{Attention}(Q, K, V) = \text{softmax}(\frac{QK^\top}{\sqrt{d_k}}+M)V,
\end{split}
\end{equation}
where $\oplus$ refers to a concatenation operation, and $M$ is given by
\begin{equation}
\begin{aligned}
{M}_{ij} &= \begin{cases} 0, &\text{allow to attend,} \\ -\infty, &\text{prevent from attending.} \end{cases} \label{eq:att:mask_encoder}
\end{aligned}
\end{equation}

\paragraph{Decoder:} suppose that $(w_{m+1}, \ldots, w_{m+l-1})$ are words generated until step $l-1$,  then the next word $w_{m+l}$ is predicted according to:
\begin{equation}\label{logits}
\small
    P(w_{m+l}|w_1,\dots,w_{m+l-1}) = \text{softmax}(W_s O(w_{m+l-1})),
\end{equation}
where $O(w_{m+l-1})$ is defined by $\text{FNN}(\text{MultiHead}(E(w_{m+l-1}),E,E)$ with $E = [E(w_1),\ldots,E(w_{m+l-1})]$ the output of the encoder, and $W_s$ is a trainable parameter.

\begin{figure*}[t]
    \centering
    \includegraphics[width=\textwidth,height=7cm]{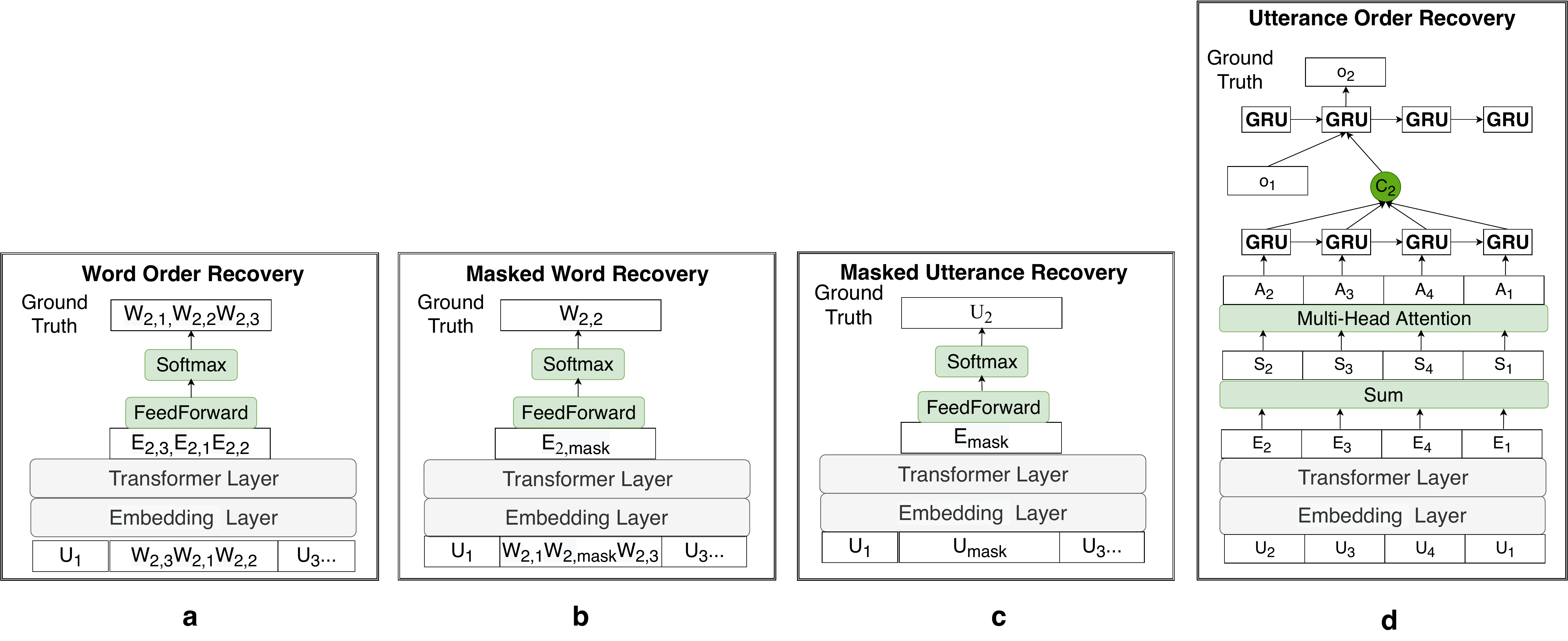}
    \caption{Auxiliary tasks.}
    \label{fig:task}
\end{figure*}

\subsection{Auxiliary Tasks}
Heading for learning the simple structure that can effectively make use of contexts for response generation, we design two kinds of auxiliary tasks including order recovery and masked content recovery. The order recovery tasks aim to enhance the capability of the self-attention module on capturing the sequential relationship among words and utterances, while the masked content recovery tasks can optimize the self-attention module to enhance semantic connection among words and utterances.  
 
\paragraph{Order recovery:}
a recent study \cite{sankar2019neural} indicates that transformer-based models are insensitive to ordering of words and utterances, which means that the information they learn could be just bag-of-words representations.  Thus, we consider recovering the correct order from random shuffling on both a word level and an utterance level to force self-attention to be aware of relative positions of words and utterances in the context.

\noindent\textbf{Word order recovery:} Figure \ref{fig:task} (a) illustrates the task. Given a randomly sampled utterance $U=(w_1,\ldots, w_k)$ from a context $\mathcal{U}$, we randomly shuffle the words in $U$ and obtain a disordered utterance $\bar{U}=(\bar{w}_1,\ldots,\bar{w}_{k})$. Then, we replace $U$ in $\mathcal{U}$ with $\bar{U}$ and form a corrupt context $\bar{\mathcal{U}}$. The goal of the task is to predict $U$ from $\bar{U}$. The loss of the task can be formulated as
\begin{equation}\label{word_order_recov}
\begin{split}
&\mathcal{L}_{\text{wor}} = -\frac{1}{k}\sum_{i=1}^k \log (p(w_i|\bar{U})),\\
&p(w_i|\bar{U}) = \text{softmax}(W_s E(\bar{w}_i)),
\end{split}
\end{equation}
where $E(\bar{w}_i)$ is obtained from $E(\bar{\mathcal{U}})$ which is the representation of $\bar{\mathcal{U}}$ given by the encoder of the generation model, $W_s$ is shared with Equation (\ref{logits}).

For this task, the mask matrix $M$ in Equation (\ref{selfatt}) is defined by:
\begin{equation} 
\small
\begin{aligned}
{M}_{ij} &= \begin{cases} 0, &\text{${w}_i$ and ${w}_j$ are in the same utterance,} \\ -\infty, &\text{${w}_i$ and ${w}_j$ are in different utterances.} \end{cases} \label{eq:att:mask_wor}
\end{aligned}
\end{equation}

\noindent\textbf{Utterance order recovery:} Figure \ref{fig:task} (d) illustrates the task. Given context $\mathcal{U}$ = $(U_1, \ldots, U_n)$, we randomly shuffle the utterances and obtain a disordered context $\bar{\mathcal{U}}$ = $(U_{o_1},\ldots,U_{o_n})$. The goal is to predict the correct positions for utterances in $\bar{\mathcal{U}}$. 
The prediction model falls in a read-process-write framework \cite{vinyals2015order}. In the reading module,  the model first represents $\bar{\mathcal{U}}$ as $\bar{E}=(\bar{E}(w_{1,1}),\ldots,\bar{E}(w_{n,m}))$ via the encoder of the generation model, where $w_{i,j}$ is the $j$-th word in utterance $U_{o_i}$ (words within an utterance are ordered), and then obtains the representation of utterance $U_{o_i}$ through
\begin{equation}
S_{i} = \sum_{j=1}^{k_i} \bar{E}(w_{i,j}),
\end{equation}
where $k_i$ is the number of words in $U_{o_i}$. $\mathcal{S}=\{S_i\}^n_{i=1}$ forms a sentence memory that is accessible by the processing module. The processing module exploits multi-head self-attention and GRU to guarantee the property that vectors retrieved from memory $\mathcal{S}$ will not change if the memory is randomly shuffled. Formally, the processing module is defined by
\begin{equation}
\begin{split}
&\{A_i\}^n_{i=1} = \text{MultiHead}(\mathcal{S},\mathcal{S},\mathcal{S}),\\
&h_t = \text{GRU}(h_{t-1}, A_t),
\end{split}
\end{equation}
where the last hidden state $h_n$ is permutation invariant regarding to input.
The writing module is another GRU that decodes $\{o_1,\ldots, o_n\}$ one by one. At step $i$, the hidden state $\bar{h}_i$ is defined by
\begin{equation}
\bar{h}_i = \text{GRU}(\bar{h}_{i-1}, [c_{i}  \oplus x_i]),
\end{equation}
where $\bar{h}_{i-1}$ is the hidden state at step $i-1$ with $\bar{h}_0=h_n$, $x_i$ is the embedding of $o_{i-1}$ (i.e., the embedding of the ground-truth position of $U_{o_{i-1}}$ in $\mathcal{U}$), and $c_i$ is a context vector which is defined via attention over $\{h_t\}_{t=1}^n$: 
\begin{equation}
\begin{split}
&c_i = \sum_{t=1}^{n} a_{i,t} h_t,\\
&\{a_{i,t}\}_{t=1}^n =\text{softmax}(\{e_{i,t}\}_{t=1}^n),\\
&e_{i,t} = V^\top \text{tanh} (W_1 \bar{h}_{i-1} + W_2 h_t +b_1),
\end{split}
\end{equation}
where $V_1$, $W_1$, $W_2$, and $b_1$ are parameters. The prediction model is finally formulated as
\begin{equation}
\begin{split}
&P(o_i|\{o_1,\ldots o_{i-1}\}, \bar{\mathcal{U}}) = \text{softmax}(u_i),\\
&u_i = \text{FNN}(\bar{h}_{i} \oplus {x}_{i} \oplus {c}_{i}).
\end{split}
\end{equation}
The loss function of the task is defined by
\begin{equation}
\mathcal{L}_{\text{uor}} = -\frac{1}{n}\sum_{i=1}^n \log (p(o_i|\{o_1,\ldots,o_{i-1}\},\bar{\mathcal{U}})).
\end{equation}
For this task and the following ones, $M$ in Equation (\ref{selfatt}) is defined as a zero matrix meaning that every pair of words can attend to each other in the context.

\paragraph{Masked content recovery:} a major challenge in context understanding is the information omission problem (e.g., coreferences) that widely exists in utterances \cite{su2019improving}. The challenge requires a model to connect semantically related words and utterances. Thus, we design masked content recovery tasks on both a word level and an utterance level to enhance the self-attention module in terms of awareness of the semantic connections. 

\begin{itemize}
\setlength\itemsep{0.01em}
\item \textbf{Word level}: for each utterance in a context, we randomly replace 15\% words with a special token [MASK].
\item \textbf{Utterance level}: we randomly pick an utterance from a context, and replace all words in the utterance with a special token [MASK]. 
\end{itemize}

Figure \ref{fig:task} (b) and Figure \ref{fig:task} (c) illustrate the task of masked word recovery (mwr) and the task of masked utterance recovery (mur) respectively. Since the only difference of the two tasks is the input, we present them in a uniform way. Given a context $\mathcal{U}=(w_1,\ldots,w_m)$, suppose that the masked context is $\bar{\mathcal{U}}=(w_1^*,\ldots,w_m^*)$, where $w_i^*=[\text{MASK}]$ if $w_i$ is masked, otherwise $w_i^*=w_i$, then, the loss of the tasks can be formulated as
\begin{equation}
\begin{split}
&\mathcal{L}_{x} = -\frac{1}{k}\sum_{i=1}^m \mathbb{I}[w_i^* \text{=} \text{[MASK]}] \log (p(w_i|\bar{\mathcal{U}})),\\
&k = \sum_{i=1}^m \mathbb{I}[w_i^* \text{=} \text{[MASK]}],\\
&p(w_i|\bar{\mathcal{U}}) =\text{softmax}(W_s E(w_i^*)),
\end{split}
\end{equation}
where $E(w_i^*)$ is the representation of $w_i^*$ obtained by passing $\bar{\mathcal{U}}$ through the encoder of the generation model, $x \in \{\text{mwr}, \text{mur}\}$ indexes the two tasks, $\mathbb{I}[\cdot]$ is an indicator function, and $W_s$ is shared with Equation (\ref{logits}). 

\subsection{Learning Objective}
The full loss function is finally defined by:
\begin{equation}\label{full_obj}
\begin{split}
\mathcal{L}_{\text{full}} &= \text{MLE} + \alpha \mathcal{L}_{\text{aux}},\\
\mathcal{L}_{\text{aux}} &= \mathcal{L}_{\text{wor}}+\mathcal{L}_{\text{uor}}+\mathcal{L}_{\text{mwr}}+\mathcal{L}_{\text{mur}},
\end{split}
\end{equation}
where $\alpha$ is a hyper-parameter as a trade-off between MLE and the objectives of the auxiliary tasks. The learning algorithm is summarized in Algorithm \ref{alg}, where $\Theta$ refers to a set of parameters including both the parameters of the generation model and the parameters of the auxiliary objectives.

\begin{algorithm}[!t]
    \caption{Optimization Algorithm}
			\SetKwData{Index}{Index}
			\SetKwInput{kwInit}{Init}
			\SetKwInput{kwOutput}{Output}
			\SetKwInput{kwInput}{Input}
			\label{alg}
    {
				
		\kwInput{Training data $\mathcal{D}$,  GlobalMaxStep $T_1$, AuxTrainEpoch $T_2$, InitialRate ${\alpha}$, BatchNumPerEpoch $N$\\}
		\kwInit{ $\Theta$} 
        \Indp
        
        \Indm
        $t=0$\\
        $\alpha=1.0$\\
        $d=\alpha/(T_2*N)$\\
        \While{$t$ $<$ $T_1$}{
            Randomly sample a mini-batch $k$ from $\mathcal{D}$.\\
            \uIf{$\alpha$ $>$ 0}{
            Compute $\mathcal{L}_{\text{aux}}$.
            }
            Compute MLE.\\
            Update the parameters of the model with respect to $\mathcal{L}_{\text{full}}$ using Adagrad.\\
            $\alpha = max(0,\alpha - d)$ \\
            $t=t+1$\\
             
        }
        \kwOutput{  $\Theta$}
    }
\end{algorithm}

%% file: experiment.tex
\section{Experiments}
We conduct experiments on DailyDialog \cite{li2017dailydialog}, PERSONA-CHAT \cite{zhang2018personalizing}, and the Ubuntu Dialogue Corpus (UDC) \cite{lowe2015ubuntu}, and compare our model with state-of-the-art baselines in terms of response quality, parameter size, and decoding speed.

\subsection{Datasets}
Both DailyDialog and PERSONA-CHAT are open domain datasets. Dialogues in DailyDialog cover a wide range of topics in daily scenarios and resemble human communications in their daily life; while PERSONA-CHAT contains multi-turn chit-chat conversations between turkers according to their assigned profiles. Since the focus of the work is how to leverage conversation history for response generation, we just append the profiles (the original ones) to the corresponding dialogues as an extension of contexts. To control the length of the dialogues and increase the number of instances, we slide a window on the training/validation/test dialogues in both datasets, and split a dialogue longer than $11$ utterances to multiple instances (i.e., the window size is $11$). Moreover, we also truncate long utterances with the first $25$ words kept. Vocabularies are formed with all words appearing in the entire data and are shared by contexts and responses. The vocabulary size of DailyDialog is $25,000$ and the vocabulary size of PERSONA-CHAT is $18,750$. The UDC data are collected from Ubuntu chat logs with two-person multi-turn conversations about Ubuntu-related problems. Here we use the same data as in \cite{zhang2019recosa}.  Table \ref{tab:datasets} reports some statistics of the three datasets.
\begin{table}[ht]
    \centering
    \small
    \scalebox{0.8}{
    \begin{tabular}{@{}llll@{}}
     \toprule
         & DailyDialog & PERSONA-CHAT & Ubuntu\\
    \midrule
        \# dialogues for training &  44,050 & 95,682 & 3980,000\\
        \# dialogues for validation &  4,176 & 11,602 & 10,000\\
        \# dialogues for test &  3,864 & 11,152 & 10,000\\
        avg. \# utter. per dialogue &7.0  & 9.4 & 4.3\\
        avg. utter. length & 13.6 &14.5 & 16.6 \\
    \bottomrule
    \end{tabular}
    }
    \caption{Statistics of the datasets.}
    \label{tab:datasets}
\end{table}

\subsection{Baselines}
We select several multi-turn response generation models as baselines: 
(1) \textbf{HRED\footnote{\url{https://github.com/hsgodhia/hred}}}: hierarchical encoder-decoder proposed in \cite{serban2016building}; (2) \textbf{VHRED\footnote{\url{https://github.com/julianser/hed-dlg-truncated}}}: an extension of HRED that factorizes response generation with latent variables \cite{serban2017hierarchical}; (3) \textbf{HRAN\footnote{\url{https://github.com/LynetteXing1991/HRAN}}}: hierarchical encoder-decoder equipped with a hierarchical attention mechanism \cite{xing2017hierarchical}; (4) \textbf{ReCoSa\footnote{\url{https://github.com/zhanghainan/ReCoSa}}}: a hierarchical transformer-based model that exhibits state-of-the-art performance on benchmarks \cite{zhang2019recosa}; and (5) \textbf{SSN}: a very recent study on enhancing dialogue generation learning with self-supervision signals extracted from utterance order \cite{wu2019self}. 
\subsection{Implementation Details}
We train the baselines and our model on RTX 2080, and initialize word embedding with GloVe vectors \cite{pennington2014glove}. In our model, the dimension of all vectors is set as $512$. The number of heads in multi-head attention is set as $8$. We adopt the Adagrad algorithm \cite{duchi2011adaptive} in optimization with a learning rate $0.05$ and a batch size $80$/$60$/$32$ in DailyDialog/PERSONA-CHAT/Ubuntu. All models are tuned on the validation sets according to perplexity. We stop training if the perplexity does not drop in three consecutive epochs. The GlobalMaxStep $T_1$ is set as 50k. The AuxTrainEpoch $T_2$ is set as 30. The BatchNumPerEpoch N is $551$/$1595$/$124,375$ for DailyDialog/PERSONA-CHAT/Ubuntu.

\begin{table*}[ht!]
    \centering
      \scriptsize
      \scalebox{1}{
        \begin{tabular}{clcccccccccccc}
            \toprule
            Dataset  & Model  & PPL & BLEU & Distinct-1 & Distinct-2 & Average & Greedy & Extrema & Parameter size & Decoding speed \\
            \midrule
            \multirow{6}{*}{DailyDialog} & HRED   & 56.22 & 0.535 & 1.553 & 3.569 & 81.393 & 65.546 & 48.109 & 34.5M & 14.79ms\\
            &HRAN   & 47.23 & 0.447 & 1.953 & 7.400 & 83.460 & 67.239 & \bf{49.599} & 38.2M & 17.15ms\\
            &VHRED & 44.79 & 0.997 & 1.299 & 6.113 & 83.866 & 67.186 & 48.570 & 34.8M & 15.67ms\\
            &SSN & 44.28 & 1.250 & 2.309 & 7.266 & 72.796 & \bf{73.069} & 44.260 & 20.0M & 12.69ms\\
            &ReCoSa  & 42.34 & 1.121 & 1.987 & 10.180 & 84.763 & 67.557 & 48.957 & 73.8M & 40.89ms\\\hline
            &Our Model      & \bf{38.60} & \bf{1.658} & \bf{3.457} & \bf{14.954} & \bf{85.224} & 69.518 & 49.069 & 20.3M/14.4M & 12.15ms\\ 
            
            \midrule
            \midrule
            
            \multirow{6}{*}{PERSON-CHAT}  & HRED   & 46.04 & 1.279 & 0.164 & 0.450 & 83.329 & 64.486 & 47.132 & 28.3M & 13.14ms\\
            &HRAN   & 41.94 & 1.997 & 0.235 & 0.771 & 82.850 & 65.556 & 47.882 & 33.1M & 18.43ms\\
            &VHRED & 42.07 & 2.181 & 0.312 & 1.915 & 82.995 & 65.578 & 46.810 & 28.8M & 20.27ms\\
            &SSN & 47.90 & 2.288 & 0.637 & 2.623 & \bf{85.002} & 66.752 & 47.461 & 15.2M & 15.82ms\\
            &ReCoSa  & 34.19 & 2.258 & 0.915 & 4.217 & 83.963 & 66.498 & 48.163 & 68.7M & 39.38ms\\\hline
            &Our Model      & \bf{33.23} & \bf{2.434} & \bf{1.279} & \bf{5.816} & 83.632 & \bf{66.778} & \bf{48.552} & 18.4M/12.5M & 13.89ms\\ 
            
            \midrule
            \midrule
            \multirow{6}{*}{Ubuntu}  & HRED   & 58.23 & 0.874 & 0.602 & 2.724 & 76.187 & 62.869 & 37.508 & 24.1M & 25.09ms\\
            &HRAN   & 48.14 & 0.922 & 0.472 & 2.217 & 76.654 & 62.145 & 37.282 & 29.5M & 31.07ms\\
            &VHRED & 52.34 & 0.906 & 0.571 & 2.933 & 76.496 & 63.051 & 36.039 & 24.7M & 30.47ms\\
            &SSN & 57.82 & \bf{1.681} & 0.557 & 2.370 & 76.431 & 61.597 & 35.976 & 12.3M & 21.11ms\\
            &ReCoSa  & 43.67 & 0.911 & 0.722 & 4.439 & 77.619 & \bf{63.239} & 36.742 & 60.6M & 45.34ms\\\hline
            &Our Model     & \bf{40.94} & 1.625 & \bf{0.783} & \bf{5.151}& \bf{78.754} & 62.738 & \bf{38.538} & 14.4M/8.5M & 22.98ms\\ 
            
            \bottomrule
        \end{tabular}
        }
        \caption{\label{tb:main_metric} Evaluation results on automatic metrics.  Numbers in bold indicate the best performing model on the corresponding metrics.}
\end{table*}

\subsection{Evaluation Metrics}
We evaluate the performance of the models in terms of response quality with both automatic metrics and human judgment. In automatic evaluation, 
besides BLEU-4 \cite{papineni2002bleu} and perplexity \cite{sutskever2014sequence}, we follow \cite{serban2017hierarchical} and employ Embedding Average (Average), Embedding Extrema (Extrema), and Embedding Greedy (Greedy) as metrics. We also follow \cite{li2015diversity} and measure the informativeness of responses with distinct-1 and distinct-2 that are calculated as the ratios of distinct unigrams and bigrams. 
In human evaluation, we randomly sample $500$ dialogues from each of the three test sets, and recruit $3$ native speakers as human annotators. For each context in the $500$ dialogues, each annotator compares a response from our model and a response from a baseline model. The two responses are top one results from greedy search, and are randomly shuffled to hide their sources. The annotators judge which response is better based on informativeness, consistency, and fluency of the responses. If an annotator cannot tell which response is better, he/she is required to label a ``tie''. Each annotator individually judges $500$ pairs for all combinations of our model and baseline models. In total, each one labels $2,500$ pairs for one dataset. Fleiss’ kappa \cite{fleiss1973equivalence} is employed to measure agreement among the annotators.  

In addition to response quality, we also compare our model with baselines on decoding speed. We calculate the average prediction time per word in response generation using all dialogues in the test sets. The efficiency comparison is conducted on a GPU environment with a single RTX 2080.

\begin{table*}[ht]
\centering
\scriptsize
\scalebox{0.9}{
\large
\begin{tabular}{lccccccc}
\multicolumn{8}{c}{DailyDialog}\\
\toprule
model variant & PPL & BLEU & distinct-1 & distinct-2 & Average & Greedy & Extrema \\ \hline
full tasks & 38.60 & 1.658 & 3.457 & 14.954 & 85.224 & 69.518 & 49.069 \\
- masked word recovery  & 38.37 & 1.365 & 2.629 & 11.135 & 85.270 & 69.901 & 49.495 \\ 
- masked utterance recovery & 39.06 & 1.407 & 2.980 & 12.544 & 85.143 & 69.667 & 49.791 \\
- word order recovery  & 41.53 & 1.082 & 2.769 & 11.166 & 85.020 & 69.417 & 49.567 \\
- utterance order recovery  & 38.69 & 1.215 & 2.551 & 9.764 & 85.253 & 69.678 & 49.644 \\
- all tasks & 46.58 & 0.903 & 1.775 & 7.136 & 84.042 & 69.017 & 48.467 \\ 
\bottomrule
\multicolumn{8}{c}{PERSONA-CHAT}\\
\toprule
model variant & PPL & BLEU & distinct-1 & distinct-2 & Average & Greedy & Extrema \\ \hline
full  tasks & 33.23 & 2.434 & 1.279 & 5.816 & 83.632 & 66.778 & 48.552 \\
- masked word recovery & 34.74 & 2.429 & 1.018 & 4.764 & 82.841 & 66.177 & 48.610 \\ 
- masked utterance recovery  & 33.49 & 2.638 & 1.045 & 5.412 & 83.402 & 66.862 & 48.810 \\
- word order recovery  & 35.06 & 2.355 & 1.028 & 4.698 & 82.503 & 66.011 & 48.350 \\
- utterance order recovery  & 33.24 & 2.484 & 1.054 & 5.011 & 82.652 & 66.025 & 47.927 \\
- all tasks & 37.16 & 1.928 & 0.938 & 4.141 & 82.104 & 65.899 & 47.162 \\ 
\bottomrule
\multicolumn{8}{c}{Ubuntu}\\
\toprule
model variant & PPL & BLEU & distinct-1 & distinct-2 & Average & Greedy & Extrema \\ \hline
full  tasks & 40.94 & 1.625 & 0.783 & 5.151 & 78.754 & 62.738 & 38.538 \\
- masked word recovery & 47.02 & 1.135 & 0.404 & 2.195 & 74.735 & 61.683 & 37.914 \\ 
- masked utterance recovery  & 42.48 & 1.543 & 0.519 & 2.419 & 76.381 & 62.203 & 37.482 \\
- word order recovery  & 48.57 & 0.962 & 0.325 & 1.537 & 77.615 & 62.819 & 38.651 \\
- utterance order recovery  & 52.04 & 1.023 & 0.359 & 1.609 & 74.982 & 59.384 & 36.825 \\
- all tasks & 57.32 & 0.851 & 0.391 & 1.765 & 73.582 & 62.581 & 37.268 \\ 
\bottomrule
\end{tabular}}
\caption{\label{tb:ablation} Results of ablation study. }
\end{table*}

\subsection{Evaluation Results}
Table \ref{tb:main_metric} reports evaluation results on automatic metrics. Our model outperforms all baseline methods on most of the metrics on all the three datasets. 
The last two columns of the tables compare different models in terms of parameter size and decoding speed.  Note that in training, the auxiliary tasks contain parameters outside the generation model. Therefore, in the column of parameter size, we report two numbers for our model with the one before ``/'' parameter size in training and the one after ``/'' parameter size of the generation model. It is remarkable that the parameter size of our model, even in training, is smaller than HRED. In spite of this, the model still outperforms ReCoSa with only $19.5\%$/$18.2\%$/$14.0\%$ parameters on the DailyDialog/PERSONA-CHAT/Ubuntu data. This is because (1) the auxiliary tasks can effectively aid the learning of the generation model in our method; and (2) ReCoSa, although in a deep structure, is still inadequate in terms of context modeling due to the RNN-based encoder and the only utterance-level attention. Besides the superior performance on response quality, our model also enjoys a fast decoding process, thanks to the small model size. In terms of decoding speed, our model is comparable with HRED, and 2x faster than ReCoSa.  The generation model of SNN is just a simple RNN sequence-to-sequence with one layer encoder and one layer decoder. Therefore, our model is comparable with SSN in terms of complexity and speed. However, SSN is worse than our model on response quality due to (1) the RNN-based seq2seq model in SSN is worse than a transformer-based structure on the benchmarks used in this work, which has been indicated by \citet{sankar2019neural}; (2) SSN only considers utterance order, while we also leverage word order, word content, and utterance content in learning. In fact, we find that the proposed auxiliary tasks can improve a 2-layer (one for encoder and one for decoder) RNN-based seq2seq model as well, as reported in Supplementary Material. On most metrics, RNN with full auxiliary tasks is better than SSN but worse than the proposed model. 


Table \ref{tb:humanEvaluation} summarizes human evaluation results.  We can see that our model outperforms all baseline models, and most of the kappa values exceed $0.6$ indicating substantial agreement among the annotators. Based on the annotation results, we find that our model tends to generate diverse and context consistent responses, indicating the effect of the auxiliary tasks.

\subsection{Discussions}
To further understand the merit of the auxiliary tasks, we make some analysis regarding to the following questions: \textbf{Q1} how do the simple architecture learned with the auxiliary tasks compare with a deep architecture; \textbf{Q2} if learning with the auxiliary tasks can also improve deep architectures; and \textbf{Q3} how different auxiliary tasks affect the performance of the model.

\begin{figure*}[!ht]

\centering
\includegraphics[width=0.33\textwidth,height=5cm]{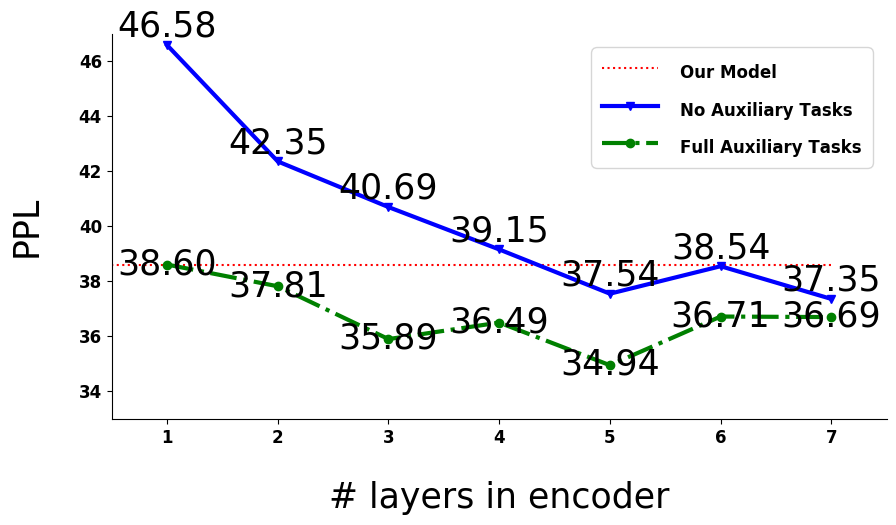}\hfill
\includegraphics[width=0.33\textwidth,height=5cm]{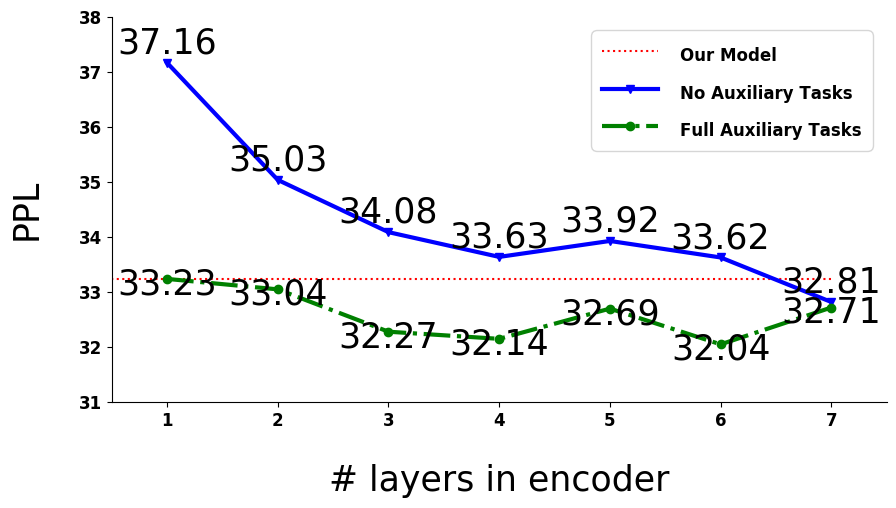}\hfill
\includegraphics[width=0.33\textwidth,height=5cm]{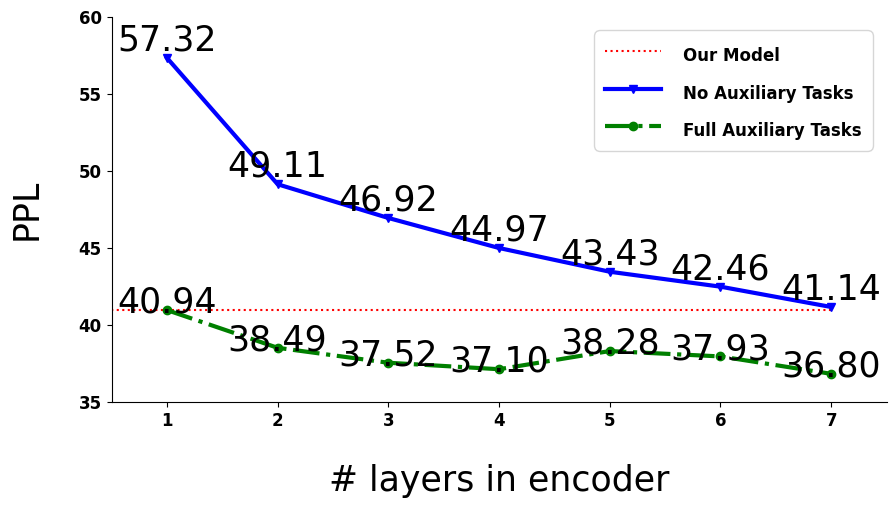}

\caption{Performance of deep architectures. (a) DailyDialog; (b) PERSONA-CHAT; (c) Ubuntu}
\label{fig:turns-trend}
\end{figure*}
\textbf{Answer to Q1:} we aim to move one step further to understand how the auxiliary tasks enhance the capability of the simple generation model on context understanding. While this is not trivial for neural models, we assume that one can let a transformer-based model capture more semantics in contexts by stacking more layers in the encoder, and examine to what extent the simple model learned with the auxiliary tasks is equivalent to a deep architecture. Figure \ref{fig:turns-trend} compares our model with deep architectures in terms of perplexity on the three datasets, in which we get the deep architectures by stacking transformer layers in the encoder of our model. The dotted lines represent our model learned with the auxiliary tasks, and the solid lines represent the deep architectures learned with MLE. Approximately, our model is equivalent to a deep model with a 4-layer encoder on the DailyDialog data, a 6-layer encoder on the PERSONA-CHAT data, and a 7-layer encoder on the Ubuntu data.

\begin{table}[!t]
\centering
\small
\scalebox{1}{
\begin{tabular}{lcccc} 
\multicolumn{5}{c}{DailyDialog}\\
\toprule
models & win  & loss  & tie  & kappa \\ \hline
Our Model v.s. HRED & 0.42 & 0.13 & 0.45 &0.675\\
Our Model v.s. VHRED & 0.38 & 0.19 & 0.43 &0.634\\ 
Our Model v.s. HRAN & 0.31 & 0.16 & 0.53 &0.587\\
Our Model v.s. SSN & 0.36 & 0.22 & 0.42 &0.638\\
Our Model v.s. ReCoSa & 0.34 & 0.22 & 0.44 &0.733\\
\bottomrule
\multicolumn{5}{c}{PERSONA-CHAT}\\
 \toprule
models & win  & loss & tie  & kappa \\ \hline
Our Model v.s. HRED & 0.45 & 0.16 & 0.39 &0.867\\
Our Model v.s. VHRED & 0.39 & 0.21 & 0.40 &0.650\\ 
Our Model v.s. HRAN & 0.36 & 0.23 & 0.41 &0.621\\
Our Model v.s. SSN & 0.49 & 0.12 & 0.39 &0.695\\
Our Model v.s. ReCoSa & 0.39 & 0.29 & 0.32 &0.566\\
\bottomrule
\multicolumn{5}{c}{Ubuntu}\\
 \toprule
models & win  & loss  & tie  & kappa \\ \hline
Our Model v.s. HRED & 0.49 & 0.14 & 0.37 &0.692\\
Our Model v.s. VHRED & 0.48 & 0.18 & 0.34 &0.603\\ 
Our Model v.s. HRAN & 0.47 & 0.13 & 0.40 &0.612\\
Our Model v.s. SSN & 0.45 & 0.18 & 0.37 &0.698\\
Our Model v.s. ReCoSa & 0.39 & 0.27 & 0.34 &0.672\\
\bottomrule
\end{tabular}}
\caption{  Human evaluation results. The ratios are calculated by combining annotations from three judges together.}
\label{tb:humanEvaluation}
\end{table}
\textbf{Answer to Q2:} since the auxiliary tasks are useful for the simple model, it is also interesting to check if they work as well for deep architectures.  Figure \ref{fig:turns-trend} shows the results, in which the dash-dotted lines represent the deep architectures learned with the full auxiliary tasks. First of all, we can conclude  that the auxiliary tasks are also useful for deep architectures, since there is clear PPL drop for the same models learned with and without (i.e., the solid lines) the auxiliary tasks. Secondly, the auxiliary tasks are more useful for simple structures, since the gap between the same models learned with and without the tasks becomes smaller and smaller when the number of encoding layers increases. The results indicate that after stacking enough layers, the effect of the auxiliary tasks is overwhelmed by the model itself. Therefore, the merit of the auxiliary tasks is to allow us to learn a generation model that enjoys both efficacy and efficiency, which is exactly the goal of the work. Improvement with respect to the number of layers of the encoder on UDC is more steady than that on DailyDialog and PERSON-CHAT. This is because the training set of UDC is much larger than those of the other two datasets.

\textbf{Answer to Q3:} we keep the architecture of the generation model and remove the objectives of the auxiliary tasks one at a time from the full learning objective given by Equation (\ref{full_obj}). Table \ref{tb:ablation} reports the ablation results. First of all, all auxiliary tasks are useful as removing any of them will cause a performance drop. When all auxiliary tasks are removed, the approach degenerates to learning a 2-layer transformer architecture 
through MLE. Without any optimization on context understanding, the simple structure is worse than ReCoSa. Secondly, on DailyDialog and UDC, order recovery tasks are more crucial than content recovery tasks due to the order insensitive nature of self-attention. 
Finally, on PERSONA-CHAT, word-level recovery tasks matter more than utterance-level recovery tasks. This might stem from the fact that in PERSONA-CHAT, dialogues highly depend on the profiles used as contexts. In many cases, utterances are just formed by copying a proportion of words from the profiles. Thus, recognizing the semantic connections and the relationship among words in contexts is more critical for the data. 

%% file: conclusion.tex
\section{Conclusions}
We propose a simple generation model with order recovery and masked content recovery as auxiliary tasks. Evaluation results on three benchmarks indicate that our model can significantly outperform state-of-the-art deep generation models in terms of both response quality and decoding speed.  

%% file: Supplementary.tex


\section{RNN with Auxiliary Tasks}
As a follow-up investigation, we are curious about if the auxiliary tasks can enhance the  performance of other simple architectures in the task of multi-turn response generation. Table \ref{tb:metric} reports the results on the three benchmarks, where the simple architecture is an RNN-base seq2seq model with one layer encoder and one layer decoder. The architecture of the model is the same as the one in SSN, that is the encoder is defined with a bi-directional GRU, the decoder is defined with a unidirectional GRU, and the decoder is equipped with an attention mechanism on the input context. From the results, we can see that the auxiliary tasks are also useful for the RNN architecture, although it is still worse than the proposed transformer-based architecture under the same learning protocol. On most metrics, the RNN model is better than SSN, since it leverages signals from full auxiliary tasks. 


\begin{table*}[th]
  \centering
  \resizebox{\linewidth}{!}{
  \begin{tabular}{c l c c c c c c c c c}
  \toprule
                                 &Model             & PPL & BLEU & Distinct-1 & Distinct-2 & Average & Greedy & Extrema  \\
        \midrule
        \multirow{3}{*}{DailyDialog}  & RNN & 47.69 & 0.668 & 1.001 & 3.563 & 80.191 & 71.211 & 45.526\\
                                 &  RNN+Auxiliary Tasks & 42.46 & 1.271 & 3.153 & 12.454 & 75.259 & 72.077 & 45.490\\
                                 &   SSN & 44.28 & 1.250 & 2.309 & 7.266 & 72.796 & 73.069 & 44.260\\
                                 &  Our Model & 38.60 & 1.658 & 3.457 & 14.954 & 85.224 & 69.518 & 49.069\\
        \midrule
        \multirow{3}{*}{PERSON-CHAT}  & RNN & 42.51 & 1.869 & 0.172 & 0.501 & 81.855 & 64.284 & 46.504\\
                                 &  RNN+Auxiliary Tasks & 38.20 & 2.356 & 0.986 & 4.037 & 84.907 & 66.951 & 48.419\\
                                 &  SSN & 47.90 & 2.288 & 0.637 & 2.623 & 85.002 & 66.752 & 47.461\\
                                 &  Our Model & 33.23 & 2.434 & 1.279 & 5.816 & 83.632 & 66.778 & 48.552\\
        \midrule
        \multirow{3}{*}{Ubuntu}  & RNN & 62.56 & 0.971 & 0.398 & 1.218 & 74.318 & 59.027 & 34.331\\
                                 &  RNN+Auxiliary Tasks & 55.44 & 1.479 & 0.602 & 3.494  & 76.494 & 62.381 & 38.139\\
                                 &  SSN & 57.82 & 1.681 & 0.557 & 2.370 & 76.431 & 61.597 & 35.976\\
                                 &  Our Model & 40.94 & 1.625 & 0.783 & 5.151 &78.754 & 62.738 & 38.538\\
        \bottomrule
  \end{tabular}
  }
  \caption{Evaluation results on automatic metrics. }
  \label{tb:metric}
\end{table*}